\documentclass{article}
\usepackage{spconf,amsmath,amssymb,graphicx}
\usepackage{graphics}
\usepackage{algorithm}
\usepackage{algpseudocode}
\usepackage{xcolor}
\usepackage{subcaption}
\usepackage{arydshln} 
\usepackage{hyperref}
\title{Exploring Diffusion Models for \\ Unsupervised Video Anomaly Detection}

\name{Anil Osman Tur$^{\star}$$^{\diamond}$ \quad Nicola Dall'Asen$^{\star\ddagger}$ \quad Cigdem Beyan$^{\star}$ \quad Elisa Ricci$^{\star}$$^{\dagger}$} 
\address{
  $^{\star}$Department of Information Engineering and Computer Science, University of Trento, Trento, Italy \\
  $^{\diamond}$Energy Efficient Embedded Digital Architectures Unit, Fondazione Bruno Kessler, Trento, Italy \\
  $^{\ddagger}$ Department of Computer Science, University of Pisa, Pisa, Italy \\
  $^{\dagger}$Deep Visual Learning Research Group, Fondazione Bruno Kessler, Trento, Italy
}

\newcommand\blfootnote[1]{%
  \begingroup
  \renewcommand\thefootnote{}\footnote{#1}%
  \addtocounter{footnote}{-1}%
  \endgroup
}

\begin{document}
\ninept
\maketitle
\begin{abstract}
This paper investigates the performance of diffusion models for video anomaly detection (VAD) within the most challenging but also the most operational scenario in which the data annotations are not used. As being sparse, diverse, contextual, and often ambiguous, detecting abnormal events precisely is a very ambitious task. To this end, we rely only on the information-rich spatio-temporal data, and the reconstruction power of the diffusion models such that a high reconstruction error is utilized to decide the abnormality. Experiments performed on two large-scale video anomaly detection datasets demonstrate the consistent improvement of the proposed method over the state-of-the-art generative models while in some cases our method achieves better scores than the more complex models. This is the first study using a diffusion model and examining its parameters' influence to present guidance for VAD in surveillance scenarios.
\blfootnote{\copyright~2023 IEEE. Personal use of this material is permitted. Permission from IEEE must be obtained for all
other uses, in any current or future media, including reprinting/republishing this material for advertising or
promotional purposes, creating new collective works, for resale or redistribution to servers or lists, or reuse
of any copyrighted component of this work in other works.}
\end{abstract}
\vspace{-0.2cm}
\section{Introduction}
\label{sec:intro}
\vspace{-0.2cm}

Automated video anomaly detection (VAD) has become an essential task in the computer vision community as a consequence of the exponential increase in the number of videos being captured. VAD is relevant to several applications in intelligent surveillance, and behavior understanding \cite{jebur2022review,sultani2018real,liu2018ano_pred,chandola2009anomaly,mohammadi2021image,beyan2013detecting},
to name a few. Anomaly is commonly defined as a rare or unexpected or unusual entity, that diverges significantly from normality, which is defined as expected and common. Despite being sparse and diverse, the abnormal events are also very contextual, and often ambiguous, thus they challenge the performance of the VAD models \cite{ren2021deep}.

Data labeling is already a notoriously expensive and time-consuming task and considering the aforementioned characteristics of the abnormal events, it is almost infeasible to collect all possible anomaly samples to perform \emph{fully-supervised} learning methods. Therefore, a typical approach in VAD, is to train a \emph{one-class} classifier that learns from the \emph{normal} training data
\cite{ravanbakhsh2017abnormal,sabokrou2017deep,zaheer2020old}.
However, the data collection problem occurring for fully-supervised learning almost remains for the one-class classifier, since it is unfeasible to have access to every variety of normal training data, given the dynamic nature of real-world applications and the wide range of normal classes \cite{chandola2009anomaly,mohammadi2021image}. In a one-class classifier setting, it is highly possible that an unseen normal event can be misclassified as abnormal since its representation is remarkably different from the representations learned from normal training data. 

The data availability problem led some researchers to define the weakly supervised VAD, which does not rely on fine-grained per-frame annotations but wields the video-level labels \cite{majhi2021dam,tian2021weakly}.
In detail, in fully-supervised VAD each \emph{individual frame} has an annotation as normal or abnormal. Instead, in weakly supervised VAD, a \emph{video} is labeled as anomalous even if only one frame of it is anomalous, and labeled as normal when all frames of it are normal. Even though performing such annotations seems relatively cheaper, it is important to notice that, in the weekly supervised setting, (a) labeling a video as normal still requires inspection of whole frames (similar to the fully-supervised setting), and (b) such methodologies often fail to localize the abnormal portion of the video, which can be impractical, e.g., when the video footage is long.

\begin{figure}[!t]
\centering
\includegraphics[width=\linewidth]{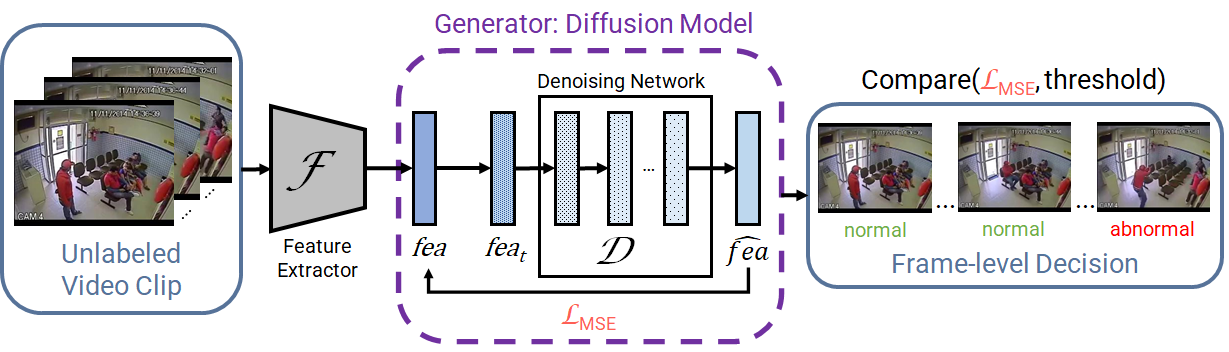}
\vspace{-0.5cm}
\caption{The proposed method takes a batch of unlabeled video clips as the input and learns to determine whether each frame is anomalous or exhibits normal behavior. Our model is a generative one, which leverages the reconstruction capability of diffusion models for unsupervised VAD. Mean Squared Error ($MSE$) distribution over a batch is used in conjunction with a data-driven threshold to decide which frames are anomalous. For the purpose of clarity, we show only one video clip.}
\vspace{-0.5cm}
\label{fig:teaser}
\end{figure}

Recently, Zaheer et al. \cite{zaheer2022generative} defined \emph{unsupervised} VAD, which takes \emph{unlabelled} videos as the input and learns to make the decision of anomaly or normality for each frame. Such a fashion is undoubtedly more challenging compared to fully, weakly, and one-class counterparts, but it literally brings in the advantage of not requiring data annotations at all. It is worth differentiating the definition of unsupervised VAD \cite{zaheer2022generative} from one-class VAD since the latter is being referred to as unsupervised in some studies \cite{gong2019memorizing,park2020learning,zaheer2020old,zaheer2022stabilizing,Schneider_2022_CVPR,li2021unsupervised}. In the case of one-class VAD, the training data distribution represents only the normality, meaning that there still exists a notion of labeling. Whereas the implementation of \emph{unsupervised} VAD \cite{zaheer2022generative} does not make any assumption regarding the distribution of the training data, and never uses the labels for model training, instead, it relies only on the spatiotemporal features of the data.

In this study, we perform \emph{unsupervised VAD} by leveraging the information-rich unlabelled videos. To do so, we only depend on the reconstruction capability of the diffusion models \cite{Karras2022edm} (see Fig. \ref{fig:teaser} for the proposed method's description). This is the first attempt that the effectiveness of the diffusion models is being investigated for VAD in surveillance scenarios. The aim of this work is to present an exploratory study: (a) to understand whether diffusion models can be effectively used for unsupervised VAD, and (b) to discover the behavior of the diffusion model \cite{Karras2022edm} in terms of several parameters of it for VAD. Experimental analysis performed on two large-scale datasets: UCF-Crime \cite{sultani2018real} and ShanghaiTech \cite{liu2018ano_pred}, demonstrate that the proposed approach always performs better than the state-of-the-art (SOTA) generative model of VAD. Moreover, in some cases, the proposed method is able to surpass more complex SOTA methods \cite{zaheer2022generative,kim2021semi}. 
% The code of our method and the SOTA \cite{zaheer2022generative} will be made publicly available \emph{upon acceptance} of this paper.
The code of our method and the SOTA~\cite{zaheer2022generative} is publicly available \href{https://github.com/AnilOsmanTur/video_anomaly_diffusion}{HERE}.

\vspace{-0.2cm}
\section{Related Work}
\label{sec:prior}
\vspace{-0.2cm}
Anomaly detection is a widely studied topic that regards several tasks such as medical diagnosis, fault detection, animal behavior understanding, and fraud detection. Interested readers can refer to a recent survey: \cite{chandola2009anomaly}. Below, our review focuses on VAD in \emph{surveillance scenarios}. We also present the definition and notations of diffusion models and state the methodology we follow for VAD.

\noindent
\textbf{Video Anomaly Detection in Surveillance Scenarios.} VAD has been typically solved as an outlier detection task (i.e., one-class classifier), in which a model is learned from the normal training data (requiring data annotations), and during testing, an abnormality is detected with the approaches such as distance-based 
\cite{ramachandra2020learning},
reconstruction-based 
\cite{ravanbakhsh2017abnormal}
or probability-based 
\cite{hinami2017joint}. 
Such approaches might result in an ineffective classifier since they exclude the abnormal classes during training. This might occur particularly when a sufficient amount of data representing each variety of the normal class cannot be used in training. An alternative approach is using \emph{unlabelled} training data without assuming any normalcy \cite{zaheer2022generative}, referred to as \emph{(fully) unsupervised VAD}. Unlike one-class classifiers, unsupervised VAD does not require data labeling and can potentially generalize well by not excluding the abnormal data from training. Zaheer et al. \cite{zaheer2022generative} proposed a Generative Cooperative Learning composed of a generator and a discriminator mutually being trained together with the negative learning paradigm. The generator which is an autoencoder reconstructs the normal and abnormal representations while using the negative learning approach to help the discriminator to estimate the probability of an instance being abnormal with a data-driven threshold. That approach \cite{zaheer2022generative} conforms that anomalies are less frequent than normal events and events are often temporally consistent. In this study, we follow the unsupervised VAD definition in \cite{zaheer2022generative}. Unlike \cite{zaheer2022generative}, our method relies only on a generative architecture, which is based on a diffusion model. We, first time in this study, investigate the effectiveness of the diffusion models for VAD in surveillance scenarios, by reporting how individual parameters affect the model performance, and by comparing them with the SOTA.

\noindent
\textbf{Diffusion Models.} Diffusion models (DMs)~\cite{sohl2015deep,ho2020denoising} are a type of generative model that gains the ability to generate diverse samples by corrupting training samples with noise and learning to reverse the process. These models have achieved SOTA performance in tasks such as text-to-image synthesis~\cite{saharia2022photorealistic},
semantic editing~\cite{meng2021sdedit},
and audio synthesis~\cite{hawthorne2022multi}.
They have also been used in representation learning for discriminative tasks like object detection~\cite{chen2022diffusiondet}, image segmentation~\cite{gu2022diffusioninst}, and disease detection~\cite{wolleb2022diffusion}. This study is the first attempt to apply DMs for video anomaly detection.

DMs are formulated as a progressive addition of Gaussian noise of standard deviation $\sigma$ to an input data point $x_T$ sampled from a distribution $p_{data}(x)$ with standard deviation $\sigma_{data}$. The noised distribution $p(x,\sigma)$, for $\sigma \gg \sigma_{data}$, becomes isotropic Gaussian and allows to sample a point $x_0 \sim \mathcal{N}(0, \sigma_{max}\mathbf{I})$. This point is gradually denoised with noise levels $\sigma_0 = \sigma_{max} > \sigma_{T-1} > \dots > \sigma_1 > \sigma_T = 0$ into new samples distributed according to the dataset distribution. DMs are trained with Denoising Score Matching~\cite{hyvarinen2005estimation}, where a denoiser function $D_\theta(x; \sigma)$ minimizes the expected $L_2$ denoising error for samples drawn from $p_{data}$ for every $\sigma$:
\vspace{-0.2cm}
\begin{equation}
\mathbb{E}_{x\sim p_{data}} \mathbb{E}_{\epsilon \sim \mathcal{N}(0, \sigma \mathbf{I)}} || D_{\theta}(x + \epsilon ; \sigma) - x ||^2_2 ,
\vspace{-0.1cm}
\end{equation}
\vspace{-0.2cm}
and the score functions used in the reverse process become:
\begin{equation}
\vspace{-0.1cm}
\nabla \log p(x;\sigma) = (D_{\theta}(x;\sigma) - x)s / \sigma^2 .
\vspace{-0.1cm}
\end{equation}
In this paper, we adapt the diffusion model of \cite{Karras2022edm}, whose details are described in the next section.

\begin{table}[b!]
\caption{ The design choice of \emph{k-diffusion}.
$T$ is the Number of Function Evaluations (NFEs) executed during sampling. The corresponding sequence of time steps is $\{t_0, t_1, \dots, t_T\}$, where $t_T = 0$.
$F_\theta$ represents the raw neural network.}
\vspace{-0.35cm}
\label{tab:specifics}
\resizebox{0.95\columnwidth}{!}{%
\begin{tabular}{lc}
\hline
\textbf{Sampling} &  \\
ODE solver & LMS \\ 
Time steps & $ ( {\sigma_{max}}^\frac{1}{\rho} + \frac{i}{T-1} ( {\sigma_{min}}^\frac{1}{\rho}-{\sigma_{max}}^\frac{1}{\rho} ))^\rho $ \\
\vspace*{-2.5mm}\\ \hline
\textbf{Network and preconditioning} &  \\
Architecture of $F_\theta$ & Any, MLP in our case \\ 
Skip scaling $c_\text{skip}(\sigma)$ & $\sigma_{data}^2 / \left(\sigma^2 + \sigma_{data}^2 \right)$ \\
Output scaling $c_\text{out}(\sigma)$ & $\sigma \cdot \sigma_{data} / \sqrt{\sigma_{data}^2 + \sigma^2}$ \\ 
Input scaling $c_\text{in}(\sigma)$ & $1 / \sqrt{\sigma^2 + \sigma_{data}^2}$ \\ 
Noise cond. $c_\text{noise}(\sigma)$ & $\frac{1}{4} \ln(\sigma)$ \\
\vspace*{-2.5mm}\\ \hline
\textbf{Training} &  \\
Noise distribution & $\ln(\sigma) \sim \mathcal{N}(P_\text{mean}^{}, P_\text{std}^2)$ \\ 
Loss weighting & $\left( \sigma^2\!+\!\sigma_{data}^2 \right) / (\sigma \cdot \sigma_{data})^2$ \\
\vspace*{-2.5mm}\\ \hline
\end{tabular}
}
\end{table}

\vspace{-0.2cm}
\section{Method}
\label{sec:method}
\vspace{-0.2cm}
Given a video clip, we first extract features from a 3D-CNN ($F$) both in training and testing. These features are supplied to the generator, which is a diffusion model, to reconstruct them without using the labels. We follow the diffusion model variant proposed in \cite{Karras2022edm} and refer to it as \textit{k-diffusion}. It disentangles the design choices of previous diffusion models and provides a framework where each component can be adjusted separately, as shown in Table~\ref{tab:specifics}. In particular, Karras et al. \cite{Karras2022edm} exposes the issue of expecting the network $D_\theta$ to work  well in high noise regimes, i.e. when $\sigma_t$ is high. To solve this, \textit{k-diffusion} proposes a $\sigma$-dependent skip connection, allowing the network to perform $x_0$ or $\epsilon$-prediction, or something in between based on the noise magnitude. The denoising network $D_\theta$, therefore, is formulated as follows:
\vspace{-0.2cm}
\begin{equation}
D_\theta(x; \sigma) = c_{\text{skip}}(\sigma) ~x + c_{\text{out}}(\sigma) ~F_\theta \big( c_{\text{in}}(\sigma) ~x; ~c_{\text{noise}}(\sigma) \big) \text{,}
\vspace{-0.1cm}
\label{eq:preconditioning}
\end{equation}
where $F_\theta$ becomes the effective network to train, $c_{skip}$ modulates the skip connection, $c_{in}(\cdot)$ and $c_{out}(\cdot)$ scale input and output magnitudes, and $c_{noise}(\cdot)$ scales $\sigma$ to become suitable as input for $F_\theta$.

\begin{algorithm}[!t]
    \caption{Anomaly detection with denoising diffusion}
      \label{alg_1}
      \begin{algorithmic}[1]
        \scriptsize
        \Require Batch of video clips $x$, feature extractor network $F$, denoising network $D$, denoising step $t$, threshold sensitivity $k$
        \State fea = $F(x)$ \# Feature extraction with the backbone $F$
        \State $\epsilon \sim \mathcal{N}(0, \textbf{I})$ \# Noise sampling for \emph{k-diffusion}
        \State $\text{fea}_{t} = \text{fea} + \epsilon * \sigma_{t}$ \# Diffusion input corruption
        \State $\hat{\text{fea}} = sampling( \text{D}(\text{fea}_{t}, \sigma_t) )$ \# Reconstruction of a feature vector with \emph{k-diffusion} algorithm
        \State $L_b = MSE(\text{fea},\hat{\text{fea}})$ \# Reconstruction loss computation
        \State $L_{th} = \mu_p + k \sigma_p$ \# Data-driven threshold $L_{th}$
        \State $\text{If } L_{th} < L_b$ \text{then abnormal, otherwise normal } \# $L_b$ is the loss in a batch
        \State $\textbf{Return } \text{Normal or Abnormal}$
    \end{algorithmic}
\end{algorithm}

Several hyperparameters control the diffusion process in \textit{k-diffusion}, and we extensively explore the role of training noise -- distributed according to a log-normal distribution with parameters $(P_{mean}, P_{std})$ -- and sampling noise with boundary values of $\sigma_{min}$ and $\sigma_{max}$. These distributions are crucial choices depending on the task and on the dataset~\cite{chen2023importance}. Given we use diffusion models on an unprecedented task and on new datasets we do not rely on parameters from literature, instead, we perform an extensive study of the correlation between noise and performance on the task in Sec.~\ref{sec:diffAnal}.

The reverse process of a DM does not need to start from noise with variance $\sigma^2_{max}$ but it can place at any arbitrary step $t \in (0, T)$, with $\sigma^2_{max} = \sigma^2_0$ as shown in~\cite{meng2021sdedit}. Given a real data point $x$, we can sample $x_t \sim \mathcal{N}(x, \sigma_t \mathbf{I})$
and then apply the reverse process to $x_T$. This allows for retaining part of the information of the original data point -- the low-frequency component -- and removing the high-frequency component. We exploit this property to remove the components associated with abnormal parts of the clip by adding Gaussian noise. Then, we measure the goodness of reconstruction using mean squared error (MSE), meaning that a high reconstruction error might indicate the presence of abnormal activity. The choice of the starting point $t$ for this procedure is a crucial hyperparameter of the method, as it controls the realism-faithfulness tradeoff as described in~\cite{meng2021sdedit}. Refer to Sec.~\ref{sec:diffAnal} presenting a study to understand the influence of this tradeoff on VAD. 

We adopt \cite{zaheer2022generative} to decide whether a video frame is anomalous. In detail, the decision for a single video frame is made by keeping the distribution of the reconstruction loss ($MSE$) of each instance over a batch. The feature vectors resulting in higher loss refer to anomalous and smaller loss refers to normal while this decision is made through a data-driven threshold ($L_{th}$), defined as $L_{th}$ $=$ $\mu_p$ $+$ $k$ $\sigma_p$ where $k$ is a constant,  $\mu_p$ and $\sigma_p$ are the mean and standard deviations of the $MSE$ loss for each batch. The anomaly detection phase is given in Algorithm \ref{alg_1}.
\vspace{-0.3cm}

\begin{table*}[!t]
\caption{Performance comparisons with the SOTA on (a) UCF-Crime \cite{sultani2018real} and (b) ShanghaiTech \cite{liu2018ano_pred} datasets. The best results are in bold. The second best results are \underline{underlined}. The full model of \cite{zaheer2022generative} includes generator, negative learning, and discriminator. $^\star$ indicates our implementation since the corresponding code is not publicly available. The results indicated with $^\diamond$ were taken from \cite{zaheer2022generative}. 
}
\label{table:compare}
\begin{subtable}[c]{0.5\textwidth}
\centering
\resizebox{0.70\linewidth}{!}{
\begin{tabular}{lcc}
\hline
 Method & Feature & AUC (\%) \\
\hline
Kim et al. \cite{kim2021semi}$^\diamond$ & 3D-ResNext 101 & 52.00 \\
 Autoencoder \cite{zaheer2022generative} & 3D-ResNext 101 & 56.32  \\
Autoencoder \cite{zaheer2022generative}$^\star$ & 3D-ResNext 101 & 56.27 \\
 Full model \cite{zaheer2022generative} & 3D-ResNext 101 & \textbf{68.17} \\
Full model \cite{zaheer2022generative}$^\star$ & 3D-ResNext 101 & 58.30 \\
Proposed w/ \cite{luo2021diffusion} & 3D-ResNext101 & 59.42 \\
Proposed & 3D-ResNext 101  & \underline{62.91} \\
\hdashline
 Autoencoder \cite{zaheer2022generative}$^\star$ & 3D-ResNet18 & 49.78 \\
 Full model \cite{zaheer2022generative}$^\star$ & 3D-ResNet18 & 56.86 \\
 Proposed w/ \cite{luo2021diffusion} & 3D-ResNet18 & \underline{60.52} \\
Proposed & 3D-ResNet18 & \textbf{65.22} \\ \hline
\end{tabular}
}
\subcaption{Results on UCF-Crime \cite{sultani2018real}}
\end{subtable}
\begin{subtable}[c]{0.5\textwidth}
\centering
\resizebox{0.70\linewidth}{!}{
\begin{tabular}{lcc}
\hline 
 Method & Feature & AUC (\%) \\ \hline
 Kim et al. \cite{kim2021semi}$^\diamond$ & 3D-ResNext 101 & 56.47 \\
 Autoencoder \cite{zaheer2022generative} & 3D-ResNext 101 & 62.73  \\
Autoencoder \cite{zaheer2022generative}$^\star$ & 3D-ResNext 101 & 62.05 \\
 Full model \cite{zaheer2022generative} & 3D-ResNext 101 & \textbf{72.41} \\
Full model \cite{zaheer2022generative}$^\star$ & 3D-ResNext 101 & 65.62 \\
Proposed w/ \cite{luo2021diffusion} & 3D-ResNext101 & 62.41 \\
Proposed & 3D-ResNext 101  & \underline{68.88} \\
\hdashline
 Autoencoder \cite{zaheer2022generative}$^\star$ & 3D-ResNet18 & 69.02 \\
 Full model \cite{zaheer2022generative}$^\star$ & 3D-ResNet18 & 71.20 \\
Proposed w/ \cite{luo2021diffusion} & 3D-ResNet18 & \underline{74.23} \\
Proposed & 3D-ResNet18 & \textbf{76.10} \\ \hline
\end{tabular}}
\subcaption{Results on ShanghaiTech \cite{liu2018ano_pred}}
\end{subtable}
\vspace{-0.7cm}
\end{table*}

\section{Experimental Analysis and Results}
\label{sec:experiments}
\vspace{-0.2cm}
As the \textbf{evaluation metric}, we use the area under the Receiver Operating Characteristic (ROC) curve (AUC), which is computed based on frame-level annotations of the test videos of the datasets, in line with the prior arts. To evaluate and compare the performance of the proposed method, experiments are conducted on two large-scale unconstrained \textbf{datasets}: UCF-Crime \cite{sultani2018real} and ShanghaiTech \cite{liu2018ano_pred}. The UCF-Crime dataset \cite{sultani2018real} is collected from various CCTV cameras, having different field-of-views. It is composed of in total 128 hours of videos with the annotation of 13 different real-world anomalous events such as road accidents, stealing, and explosion. We use the standard training (810 abnormal and 800 normal videos, without using the labels) and testing (130 abnormal and 150 normal videos) splits of the dataset to provide fair comparisons with SOTA. ShanghaiTech dataset \cite{liu2018ano_pred} is captured in 13 different camera angles with complex lighting conditions. We use the training split that contains 63 abnormal and 174 normal videos and the testing split composed of 44 abnormal and 154 normal videos curated in line with SOTA.

We use 3D-ResNext101 and 3D-ResNet18 as feature extractors $F$ due to their popularity in VAD \cite{chandola2009anomaly,mohammadi2021image,zaheer2022generative}.  3D-ResNext101 has a dimensionality of 2048, and 3D-ResNet18 has 512 dimensions. The denoising network $D$ is an MLP with an encoder-decoder structure. The encoder is composed of 3 layers of size \{1024, 512, 256\} while the decoder has the structure of \{256, 512, 1024\}. The learning rate scheduler and EMA of the model are taken as the default values of \textit{k-diffusion}, with an initial learning rate of $2\times10^{-4}$ and InverseLR scheduling; weight decay is set at $1\times10^{-4}$.
Segment size for feature extraction is set to 16 non-overlapping frames and the training is performed up to 50 epochs with a batch size of 8192 in line with \cite{zaheer2022generative}. The timestep $\sigma_t$ is transformed via Fourier embedding and integrated into the network through FiLM layers \cite{perez2018film} both in the encoder and the decoder segments of the network.
The hyperparameters (e.g., $P_{mean}$, $P_{std}$, $t$) used to realize \emph{k-diffusion} are given in Sec. \ref{sec:diffAnal}.

\vspace{-0.2cm}
\subsection{Comparisons with State-Of-The-Art (SOTA)}
\label{sec:sota}
\vspace{-0.2cm}
The performance of the proposed method is compared with the SOTA \cite{kim2021semi,zaheer2022generative} in Table \ref{table:compare}. Kim et al. \cite{kim2021semi} proposed a one-class VAD method, which was then adopted to perform \emph{unsupervised} VAD in \cite{zaheer2022generative}. In our comparisons, we use the unsupervised version of \cite{kim2021semi}. The proposed approach surpasses \cite{kim2021semi} within a large margin of: 10.91-12.41\% AUC. 
The comparisons between the proposed method and the autoencoder of \cite{zaheer2022generative} demonstrate that as a generative model, the proposed method is more favorable than \cite{zaheer2022generative} by better performing VAD within a margin of: 6.15-14.44\% AUC. When the features extracted from 3D-ResNext101 are used, the full model of \cite{zaheer2022generative} achieves better results than the proposed method. This is rather not surprising given that the full model of \cite{zaheer2022generative} is more complex than a generative model (i.e., autoencoder or diffusion model) since it additionally includes a discriminator and a negative learning component. Importantly, when 3D-ResNet18 is used as the backbone, the proposed method exceeds the full model of \cite{zaheer2022generative} within a large margin of: 4.9-8.36\% AUC. Such results confirm the remarkable effectiveness of \emph{k-diffusion} to perform VAD.

\vspace{-0.2cm}
\subsection{Diffusion Model Analysis}
\label{sec:diffAnal}
\begin{figure}[!t]
\centering
\includegraphics[width=0.85\linewidth]{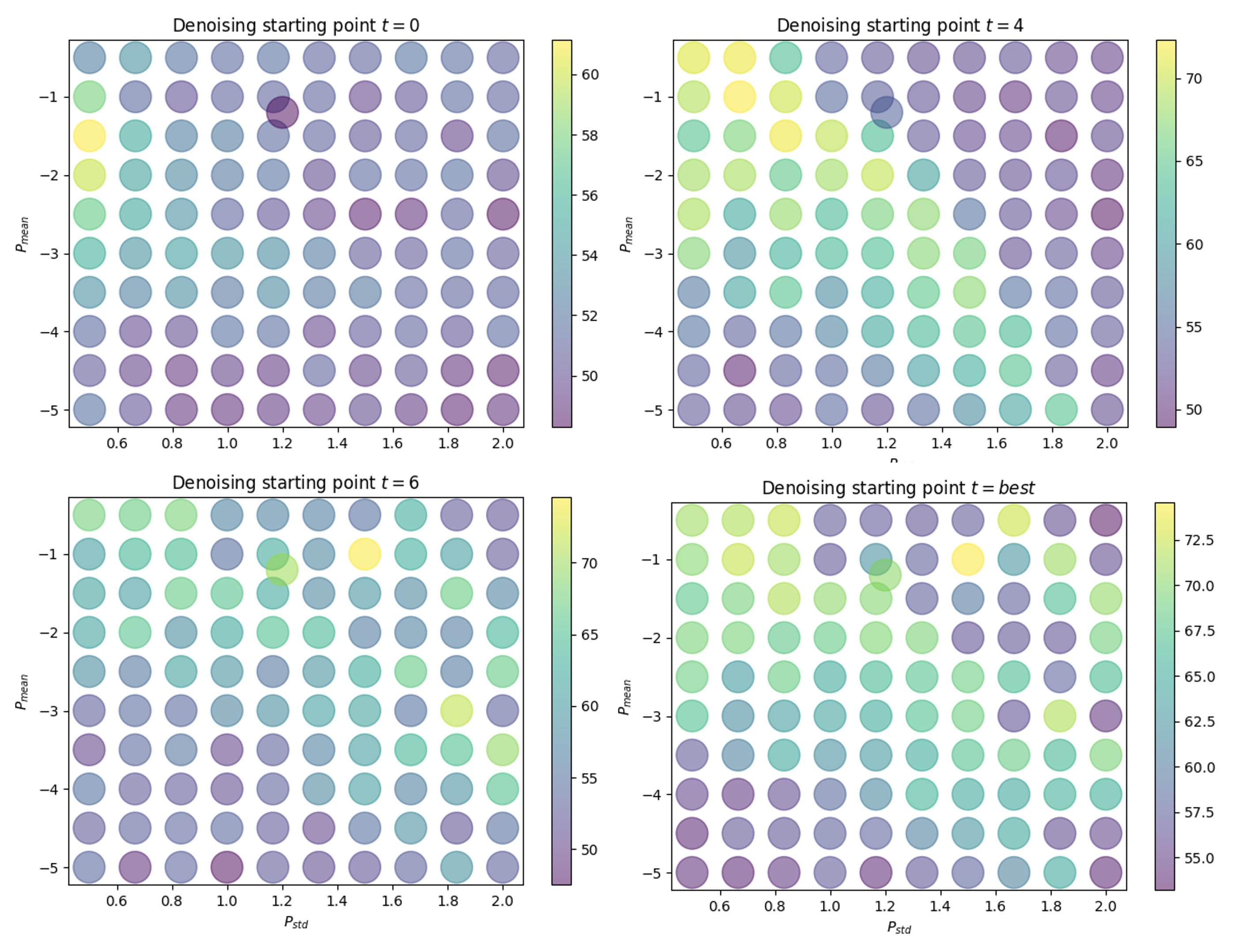}
\vspace{-0.3cm}
\caption{The effect of noise and the starting point of the reverse process while performing \emph{k-diffusion} for VAD. The results (AUC \%) belong to ShanghaiTech dataset \cite{liu2018ano_pred} with 3D-ResNet18 backbone.}
\vspace{-0.6cm}
\label{fig:noiseFig}
\end{figure}

\vspace{-0.2cm}
Below, the effect of different hyperparameters of \emph{k-diffusion} model and a comparative study regarding timestep embeddings are given. \\
\textbf{Noise.} The training and sampling noise distributions are not independent in \emph{k-diffusion} model, and we computed the relation between $(P_{mean}, P_{std})$ and $(\sigma_{min}, \sigma_{max}$) to be governed by the following formula: $\sigma_{max}, \sigma_{min} = e^{P_{mean} \pm 5 P_{std}}$. This allows us to restrict our search to two parameters instead of four. We also extracted the formula using the default parameters of \textit{k-diffusion}: $P_{mean} = -1.2, P_{std}=1.2, \sigma_{min} = 0.02$ and $\sigma_{max} = 80$. The corresponding results are given in Fig. \ref{fig:noiseFig} when the $k$ of $L_{th}$ is taken as 1 for ShanghaiTech dataset \cite{liu2018ano_pred} with 3D-ResNet18.  One can observe that, in general, a smaller value of $P_{mean}$ leads to higher results. This shows that we perform diffusion in a latent space that is well-behaved, and therefore a lower amount of noise is needed to reach an isotropic Gaussian distribution. \\
\textbf{Starting point of the reverse process.} Similar to SDEdit~\cite{meng2021sdedit} and their realism-faithfulness tradeoff, we explore the effect of different $t$ as the starting point of the reverse process. Recall that $\sigma_t > \sigma_{t+1}$ means that a $t$ close to zero indicates a noised $x_t$ closer to isotropic Gaussian, instead for $t$ closer to $T$ means, the features used are closer to the original data distribution. We target to find the best value of $t$ such that sufficient information about the structure of the clip is retained while the information about the possible anomaly is destroyed. In this way, one can obtain a higher reconstruction error, leading to deciding the associated video frame as anomalous.
The corresponding results are given in Fig. \ref{fig:noiseFig} when $k$ of $L_{th}$ is 1. $t= best$ refers to the best results obtained from $t=0$ to $t=9$, given a fixed $P_{mean}, P_{std}$ combination.
For ShanghaiTech with 3D-ResNet18 backbone, the majority of the time the starting point $t=4$ results in the best performances. The best of all results was observed when $t=6$. For all other datasets, and backbone combinations, the best results were obtained with $t=9$. Overall, increasing the $t$ value for a fixed combination of $P_{mean}, P_{std}$ improves the VAD results.
\\
\textbf{Threshold $L_{th}$.} Given the abnormality threshold $L_{th}$ $=$ $\mu_p$ $+$ $k$ $\sigma_p$, the effect of $k$ was investigated by setting its value to: 0.1, 0.3, 0.5, 0.7, and 1. For 3D-ResNext101, the best results correspond to $k$=0.5 for both ShanghaiTech and UCF Crime. For 3D-ResNet18, $k$=0.7 and 0.1 for ShanghaiTech and UCF Crime, respectively, result in the best scores. The difference between the highest and lowest performances upon changing the value of $k$ is up to 3\% AUC when the values of all other hyperparameters are kept the same.
\\
\textbf{Timestep embeddings.} As mentioned before, our method includes transforming the timestep $\sigma_t$ via Fourier embedding and integrating it into the network through FiLM layers \cite{perez2018film}. We also adopted the implementation of \cite{luo2021diffusion}, which concatenates the timestep embeddings together with its sinus and cosinus values (shown as Proposed w/ \cite{luo2021diffusion} in Table \ref{table:compare}). The results confirm the better performance of our proposal \emph{wrt} to adapting \cite{luo2021diffusion} for all cases while both surpassing the SOTA with the 3D-ResNet18 features.

\vspace{-0.3cm}

\begin{figure}[!t]
\centering
\includegraphics[width=0.75\linewidth]{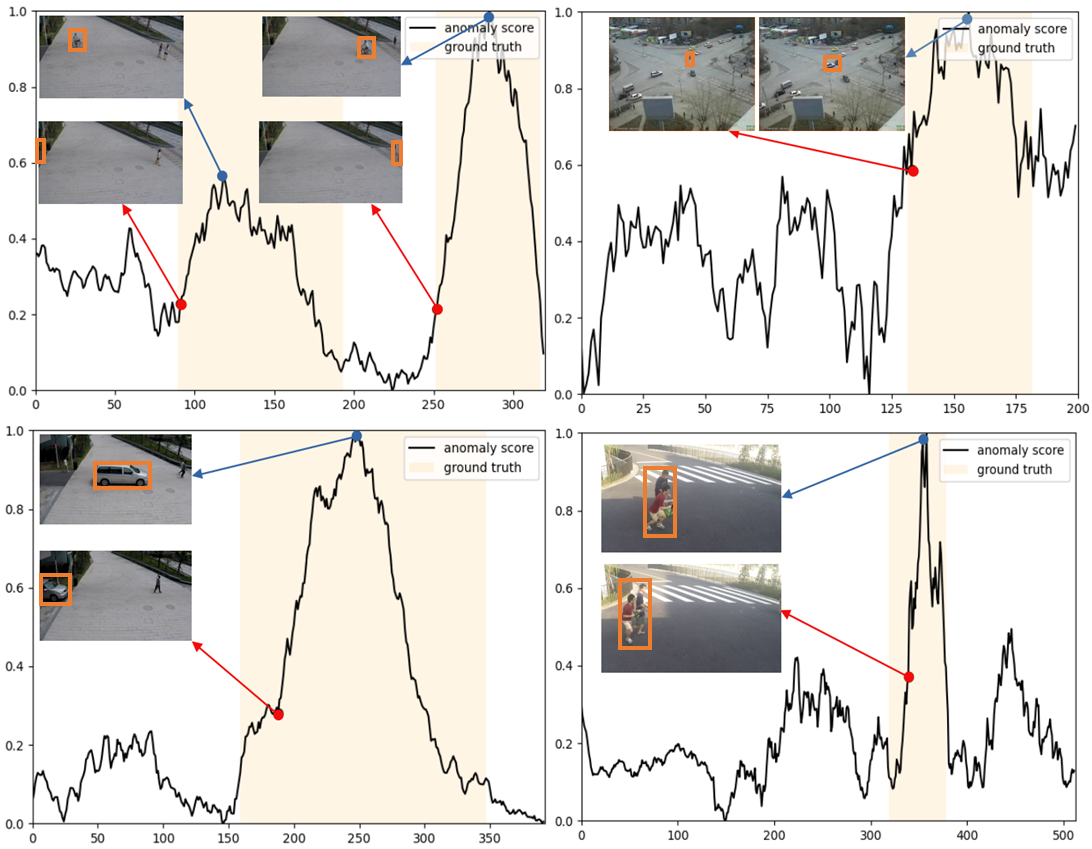}
\vspace{-0.2cm}
\caption{Predicted frame-level anomaly scores (black), predicted starting frames of an anomaly interval (red arrow), predicted highest anomaly score in an interval (blue arrow), ground-truth anomalies (yellow shadows), spatial reasoning of the anomaly (orange boxes).}
\vspace{-0.6cm}
\label{fig:anomalyResults}
\end{figure}

%\vspace{-0.1cm}
\subsection{Qualitative Results}
\label{sec:qualitative}
\vspace{-0.2cm}
Fig. \ref{fig:anomalyResults} shows the anomaly scores produced by our approach for example video clips. As seen, independent of the type of anomaly, the anomaly scores increase immediately when ground-truth anomalies start and decrease right after the ground-truth anomalies finish, showing that the proposed method is favorable for VAD.

\vspace{-0.5cm}

\section{Conclusion}
\vspace{-0.2cm}
Unsupervised video anomaly detection (VAD) presents the advantage of not requiring data annotation for learning. This solves the problems posed by the heterogeneity of normal and anomalous instances and the scarcity of anomalous data.
This paper is the first attempt to investigate the capability of diffusion models for VAD in video surveillance in which we have
specifically investigated the use of high reconstruction error as an indicator of abnormality.
The experiments performed on popular benchmarks show that the proposed model achieves better performance compared to SOTA generative model: autoencoders independent of the feature extractor used. Our model, although relying only on the reconstruction of the spatial-temporal data, is able, in some cases, to surpass the performance of more complex methods, e.g. the ones performing collaborative learning of generative and discriminative networks. We have also presented a guideline on how the diffusion models (particularly the \emph{k-diffusion} \cite{Karras2022edm} formulation) should be utilized in terms of its several parameters for VAD. The future work includes investigating the generalization ability of our method in cross-dataset settings. %The proposed method will be further explored in the tiny machine learning framework to ensure that it functions with near real-time capabilities and is potentially operable on edge devices, which is critical for their practical implementation.

\vspace{-0.3cm}

\section{Acknowledgment}
We acknowledge the support of the MUR PNRR project FAIR - Future AI Research (PE00000013) funded by the NextGenerationEU. E.R. is partially supported by the PRECRISIS, funded by the EU Internal Security Fund (ISFP-2022-TFI-AG-PROTECT-02-101100539). The work was
carried out in the Vision and Learning joint laboratory of FBK and UNITN.

\bibliographystyle{IEEEbib}
\bibliography{refs}

\end{document}